%% file: root.tex
\documentclass[letterpaper, 10 pt, conference]{ieeeconf}  

\IEEEoverridecommandlockouts                              

\usepackage{epsfig} 
\usepackage{times} 
\usepackage{amsmath} 
\usepackage{amssymb}  
\usepackage{booktabs}
\usepackage{subcaption}
\usepackage[super]{nth}
\usepackage[pagebackref,breaklinks,colorlinks]{hyperref}

\title{\LARGE \bf
LPFormer: LiDAR Pose Estimation Transformer with Multi-Task Network
}

\author{Dongqiangzi~Ye$^{\dagger1}$, Yufei~Xie$^{\dagger1}$, Weijia~Chen$^{\dagger1}$, Zixiang~Zhou$^{\dagger2}$, Lingting Ge$^{1}$ and Hassan~Foroosh$^{2}$
\\ $^{1}$ TuSimple, $^{2}$ University of Central Florida
\thanks{$^{1}$ Previous work done at TuSimple.}
\thanks{$\dagger$ Contributed equally.}
}

\newcommand{\kptr}{LPFormer}

\begin{document}

\maketitle
\thispagestyle{empty}
\pagestyle{empty}

\begin{abstract}
Due to the difficulty of acquiring large-scale 3D human keypoint annotation, previous methods for 3D human pose estimation (HPE) have often relied on 2D image features and sequential 2D annotations. Furthermore, the training of these networks typically assumes the prediction of a human bounding box and the accurate alignment of 3D point clouds with 2D images, making direct application in real-world scenarios challenging. In this paper, we present the \nth{1} framework for end-to-end 3D human pose estimation, named \kptr, which uses only LiDAR as its input along with its corresponding 3D annotations. \kptr~ consists of two stages: firstly, it identifies the human bounding box and extracts multi-level feature representations, and secondly, it utilizes a transformer-based network to predict human keypoints based on these features. Our method demonstrates that 3D HPE can be seamlessly integrated into a strong LiDAR perception network and benefit from the features extracted by the network. Experimental results on the Waymo Open Dataset demonstrate the state-of-the-art performance, and improvements even compared to previous multi-modal solutions.

\end{abstract}

\input{intro}

\input{relate}
\input{method}
\input{experiment}

\section{Conclusion}
Previous approaches to LiDAR-based 3D human pose estimation typically treated it as a stand-alone problem, overlooking the potential of leveraging the valuable features learned within other LiDAR perception networks. In this paper, we introduced the \nth{1} LiDAR-only end-to-end solution for 3D human pose estimation. Our proposed \kptr~not only achieves state-of-the-art performance on the large-scale Waymo Open Dataset, but more importantly, shows that 3D human pose estimation can be solved using only a 3D point cloud as the input and a limited amount of 3D annotations. \kptr~also proves that major LiDAR perception tasks can learn robust feature representations that can benefit other fine-grind tasks like human pose estimation. As for the future work, we plan to further enhance our \kptr~method through broad integration and fusion of LiDAR and camera data, in addition to exploring 2D weak supervision.

{\small
\bibliographystyle{IEEEtran}
\bibliography{egbib}
}

\end{document}

%% file: intro.tex
\section{Introduction}

Human pose estimation (HPE) has gained significant popularity in the image and video domain due to its wide range of applications. However, pose estimation using 3D data, such as LiDAR point cloud, has received less attention due to the difficulty associated with acquiring accurate 3D annotations. As a result, previous methods~\cite{zheng2022multi,cong2022weakly,zanfir2023hum3dil} on LiDAR-based HPE commonly rely on weakly-supervised approaches that utilize 2D annotations. These approaches often assume precise calibration between the camera and the LiDAR sensor. However, in real-world scenarios, small errors in annotations or calibration can propagate into significant errors in 3D space, thereby affecting the training of the network. Additionally, due to variations in the perspective view, it is difficult to accurately recover important visibility information by simply lifting 2D annotations to the 3D space.

In image-based HPE, the dominant approach is the top-down method~\cite{sun2019deep}, which involves first detecting the human bounding box and then predicting the single-person pose based on the cropped features. However, a significant gap exists in the backbone network between the 2D detector and the 3D detector. Most LiDAR object detectors utilize projected bird's-eye view (BEV) features to detect objects, which helps reduce computational costs. This procedure leads to the loss of separable features in the height dimension that are crucial for human pose estimation. An effective use of learned object features for HPE is still unexplored.

\begin{figure}[t]
    \begin{center}
    \includegraphics[width=\linewidth]{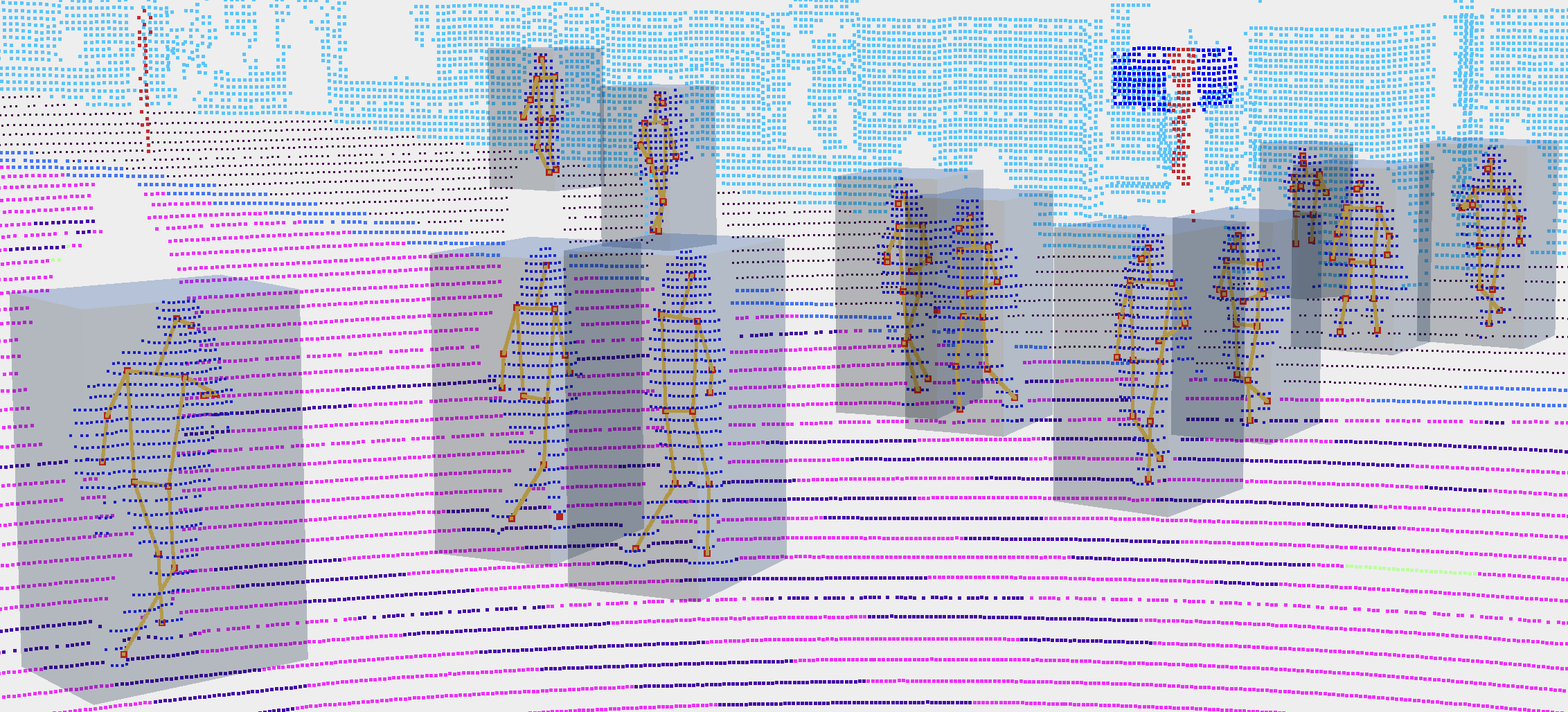}
    \end{center}
    \caption{Our method can predict 3D keypoints (red points with yellow wireframes), 3D bounding boxes, and 3D semantic segmentation in a single framework.
    }
    \label{fig:opening}
\end{figure}

\begin{figure*}[t]
  \begin{center}
    \includegraphics[width=1.0\textwidth]{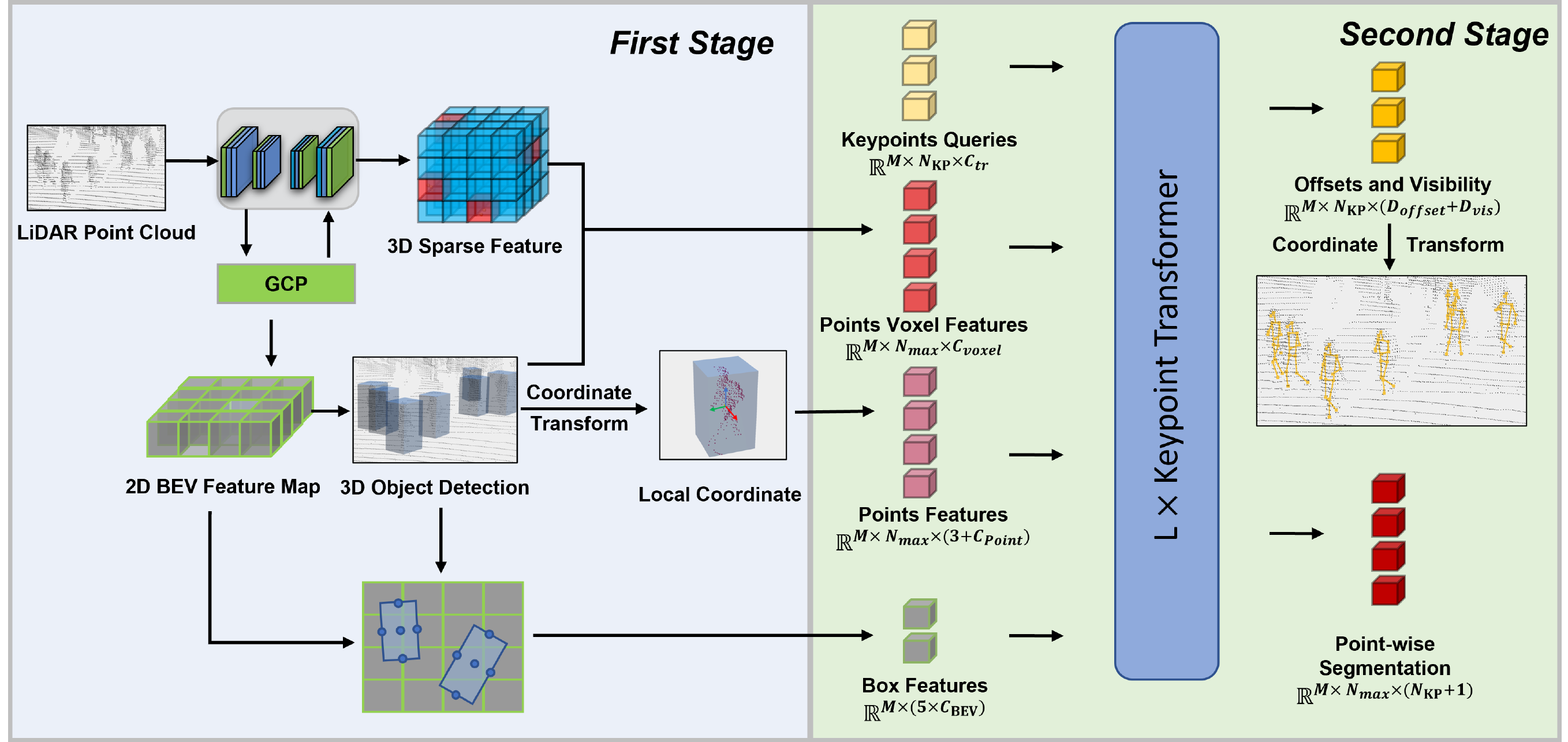}
  \end{center}
  \caption{\textbf{Main Architecture of \kptr.} Our network aims to estimate the 3D human pose for the entire frame based on the LiDAR-only input. It is comprised of two main components. The left part (blue) represents our powerful multi-task network, LidarMultiNet \cite{ye2022lidarmultinet}, which generates accurate 3D object detection and provides rich voxel and bird's-eye-view (BEV) features. The right part (green) corresponds to our Keypoint Transformer (KPTR), predicting the 3D keypoints of each human box using various inputs from our first-stage network.}
  \label{fig:architecture}
\end{figure*}

In this paper, we present \textbf{\kptr}, a complete two-stage top-down 3D human pose estimation framework that uses only LiDAR point cloud as input and is trained solely on 3D annotations. In the first stage, we adopt the design of the previous state-of-the-art LiDAR multitask network~\cite{ye2022lidarmultinet,zhou2023lidarformer} that can accurately predict human object bounding boxes while generating fine-grained voxel features at a smaller scale. The second stage extracts point-level, voxel-level, and object-level features from the point cloud inside each predicted bounding box and regresses the keypoints in a light-weighted keypoint transformer network. Our approach demonstrates that complex HPE tasks can be seamlessly integrated into the LiDAR multi-task learning framework (as shown in Figure~\ref{fig:opening}), achieving state-of-the-art performance without the need for image features or annotations. The main contributions of this paper are summarized as follows:

\begin{itemize}
\item We present the \nth{1} end-to-end 3D HPE network that only depends on LiDAR input and 3D annotations. 
\item Our proposed keypoint transformer can be integrated into a unified LiDAR perception network, extending a previously-proposed strong multitask framework to HPE tasks with only minimal computational overhead.
\item Our experimental results on large-scale datasets show state-of-the-art performance. Notably, \kptr~ currently reaches the \nth{1} place performance on the Waymo Open Dataset Pose Estimation leaderboard. Our approach also outperforms all previous camera-based and multi-modal methods on the validation set.
\end{itemize}

%% file: relate.tex
\section{Related Work}

\noindent\textbf{Image-based 3D human pose estimation}
3D human pose estimation (HPE) has been extensively studied based solely on camera images, where the human pose is represented as a parameter mesh model such as SMPL~\cite{loper2015smpl} or skeleton-based keypoints. Previous works in this area can be generally categorized into two main approaches: top-down~\cite{sun2019deep,li2022cliff} or bottom-up methods~\cite{cheng2020higherhrnet,sun2021monocular}. Top-down methods decouple the pose estimation problem to individual human detection using an off-the-shelf object detection network and single-person pose estimation on the cropped object region. In contrast, bottom-up methods first estimate the instance-agnostic keypoints and then group them together~\cite{cheng2020higherhrnet} or directly regress the joint parameters using center-based feature representation~\cite{sun2021monocular}. Some recent works~\cite{shi2022end,yang2023explicit} explored using the transformer decoder to estimate human pose in an end-to-end fashion following the set matching design in DETR~\cite{carion2020end}. However, image-based 3D HPE suffers from inaccuracies and is considered not applicable to larger-scale outdoor scenes due to frequent occlusion and the difficulty of depth estimation.

\noindent\textbf{LiDAR-based 3D human pose estimation}
To solve the depth ambiguity problem, some researchers~\cite{zimmermann20183d,xiong2019a2j} explored using depth images for the 3D HPE. Compared to the depth image, LiDAR point cloud has a larger range and is particularly applicable to outdoor scenes, such as in autonomous driving applications. Waymo Open Dataset~\cite{sun2020scalability} recently released the human keypoint annotations on both associated 2D images and 3D LiDAR point cloud. However, due to the lack of enough 3D annotation, previous works~\cite{zheng2022multi,zanfir2023hum3dil} have focused on semi-supervised learning approaches. These approaches lift 2D annotation to the 3D space and rely on the fusion of image and LiDAR features for the HPE task. 

%% file: method.tex
\section{Method}

\begin{figure*}[t]
  \begin{center}
    \includegraphics[width=0.9\textwidth]{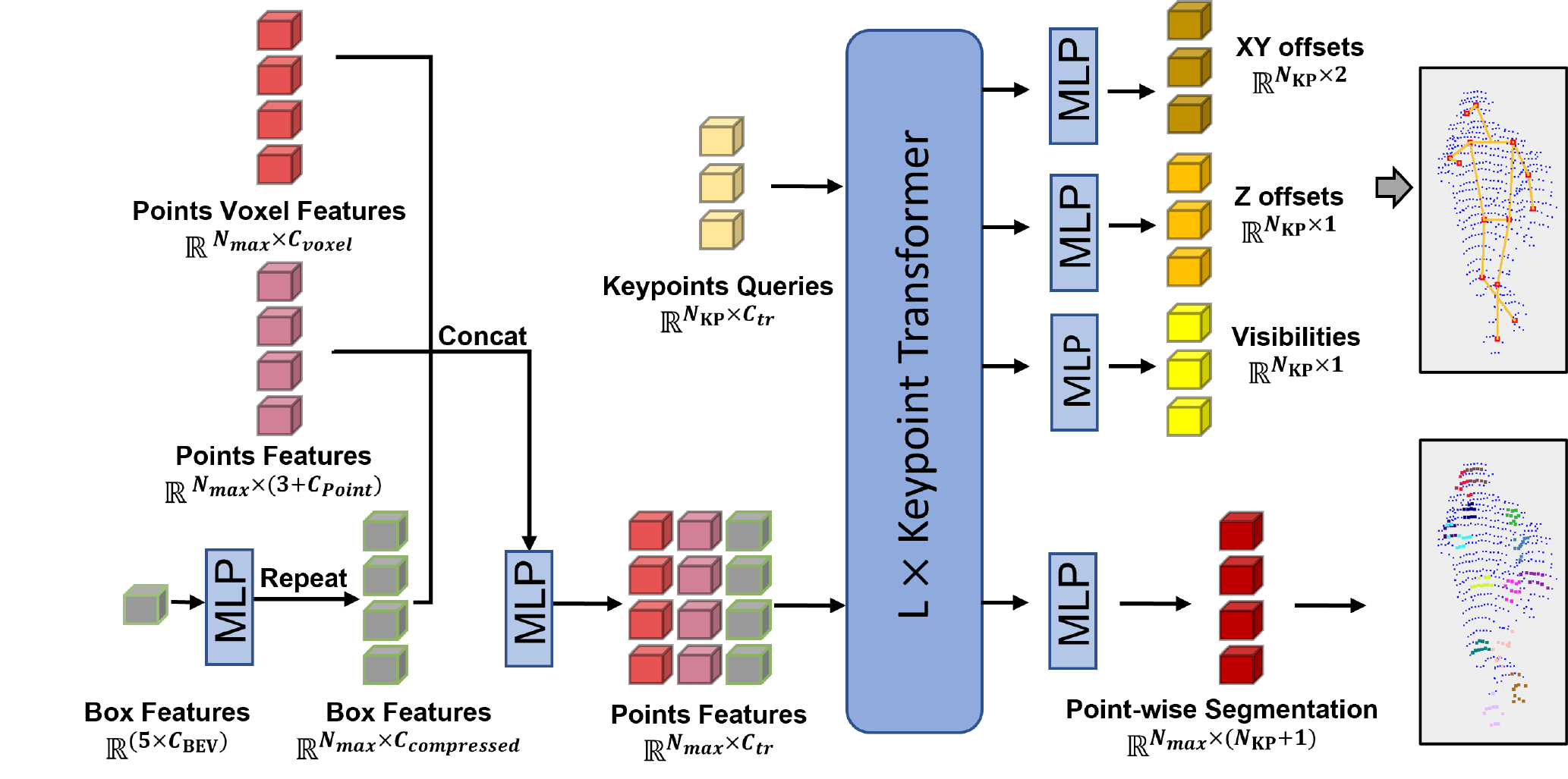}
  \end{center}
  \caption{\textbf{Illustration of Keypoint Transformer (KPTR).} In the initial stage of our KPTR, we start by compressing the feature dimension of the box features. These compressed box features are then repeated and concatenated with the point features and point voxel features. The keypoint queries are generated from learnable embedding features. Then $L$ sequences of KPTR operations are performed on the keypoint queries and point tokens. Finally, the keypoint queries are passed through three distinct MLPs to learn the XY offsets, the Z offsets, and the visibilities of the 3D keypoints. Simultaneously, the point tokens are processed by an MLP to learn the point-wise segmentation labels for the 3D keypoints, which serves as an auxiliary task.}
  \label{fig:architecture_2nd}
\end{figure*}

\kptr~is a two-stage LiDAR-only model designed for 3D pose estimation. Figure~\ref{fig:architecture} provides an overview of our framework. The input to \kptr~consists only of point clouds, represented as a set of LiDAR points $P=\{p_{i}|p_{i} \in \mathbb{R}^{3+C_{point}} \}_{i=1}^{N}$, where $N$ denotes the number of points and $C_{point}$ includes additional features such as intensity, elongation, and timestamp for each point. In the first stage, we employ a powerful multi-task network \cite{ye2022lidarmultinet} that accurately predicts 3D object detection and 3D semantic segmentation, incorporating meaningful semantic features. Inspired by a recent work \cite{zanfir2023hum3dil}, our second stage leverages a transformer-based model. This model takes various outputs from the first stage as inputs and generates 3D human keypoints $Y_{kp} \in \mathbb{R}^{N_{kp} \times 3}$ along with their corresponding visibilities $Y_{vis} \in \mathbb{R}^{N_{kp}}$, where $N_{kp}$ is the number of 3D keypoints.

\subsection{First Stage of Detection}

The first stage of our \kptr~adopts the methodology of LidarMultiNet \cite{ye2022lidarmultinet} for extracting point cloud features from raw point clouds $P$. Illustrated in Figure~\ref{fig:architecture}, it consists of a 3D encoder-decoder structure with a Global Context Pooling (GCP) module in between. The 3D object bounding boxes are predicted through the 3D detection head, which is attached to the dense 2D BEV feature map.

\textbf{Enriching point features with multi-level feature embedding}
 Within each detected bounding box, the points are transformed by a local coordinate transformation involving translation and rotation. The transformed points are then concatenated with their corresponding original point features, resulting in $P_{point} \in\mathbb{R}^{M\times N_{max} \times (3+C_{point})}$, where $M$ is the number of bounding boxes and $N_{max}$ represents the maximum number of 3D points within each bounding box. For each box, we randomly remove extra points, and pad with zero if the number of points within a box is less than $N_{max}$. Additionally, we generate point voxel features $P_{voxel} \in \mathbb{R}^{M\times N_{max} \times C_{voxel}}$ by gathering the 3D sparse features from the decoder using their corresponding voxelization index, where $C_{voxel}$ denotes the channel size of the last stage of the decoder. Similar to CenterPoint \cite{yin2021center}, for each bounding box, we adopt the BEV features at its center as well as the centers of its edges in the 2D BEV feature map as the box features $B \in \mathbb{R}^{M\times (5 \times C_{BEV})}$

\subsection{Second Stage of Keypoint Transformer}
By leveraging the capabilities of the robust first stage model LidarMultiNet \cite{ye2022lidarmultinet}, our second stage is able to exploit valuable semantic features for capturing intricate object details, including human 3D pose. Different from LidarMultiNet \cite{ye2022lidarmultinet}, we choose a transformer architecture instead of a PointNet-like \cite{qi2017pointnet} structure as our second stage, in order to effectively understand 3D keypoints by leveraging local points information through an attention mechanism. The details of our second stage are shown in Figure \ref{fig:architecture_2nd}.

Specifically, our second stage takes various features from local point features $P_{point}$, semantic voxel-wise point features $P_{voxel}$, and box-wise features $B$ to predict 3D keypoints for each pedestrian or cyclist box. Starting with a box-wise feature $B$, we first employ a multilayer perceptron (MLP) to compress its dimensions from $\mathbb{R}^{5 \times C_{BEV}}$ to $\mathbb{R}^{C_{compressed}}$. This compressed box-wise feature is then replicated as $P_{box} \in \mathbb{R}^{N_{max} \times C_{compressed}}$ and combined with point-wise features $P_{point}$ and $P_{voxel}$, resulting in $P_{cat} \in \mathbb{R}^{N_{max} \times (3 + C_{point} + C_{voxel} + C_{compressed})}$. The fused point-wise features are subjected to a simple matrix multiplication, yielding $X_{point} \in \mathbb{R}^{N_{max} \times C_{tr}}$, which serves as one part of the input for Keypoint Transformer (KPTR). The other input for KPTR is a learnable 3D keypoints query $X_{kp} \in \mathbb{R}^{N_{kp} \times C_{tr}}$. Subsequently, we employ KPTR, which consists of $L$ blocks of a multi-head self-attention and a feed-forward network, to learn internal features $X_{point}^{'}$ and $X_{kp}^{'}$. Finally,  the keypoints' internal features $X_{kp}^{'}$ are fed into three separate MLPs to predict 3D keypoints offsets along the X and Y axes $\hat{Y}_{xy} \in \mathbb{R}^{N_{kp} \times 2}$, 3D keypoints offsets along the Z axis $\hat{Y}_{z} \in \mathbb{R}^{N_{kp} \times 1}$, and 3D keypoints visibilities $\hat{Y}_{vis} \in \mathbb{R}^{N_{kp}}$. Furthermore, the point-wise internal features $X_{point}^{'}$ are processed by an MLP to estimate point-wise keypoint segmentation $\hat{Y}_{kpseg} \in \mathbb{R}^{N_{max} \times (N_{kp} + 1)}$.

For the final predictions, we combine the predicted 3D keypoints offsets $\hat{Y}_{xy}$, $\hat{Y}_{z}$, and the predicted 3D keypoints visibilities $\hat{Y}_{vis}$ to generate the human pose for each bounding box. Then we apply a reverse coordinate transformation to convert the predicted human pose from the local coordinate system to the global LiDAR coordinate system. Moreover, the predicted point-wise keypoint segmentation $\hat{Y}_{kpseg}$ serves as an auxiliary task, aiding KPTR in learning point-wise local information and enhancing the regression of 3D keypoints through the attention mechanism. In the experiments section, we will demonstrate how this auxiliary task significantly enhances the overall performance of the model.

\subsection{Training and Losses}
During the training phase, we replace the predicted bounding boxes with ground truth bounding boxes that include 3D keypoints labels. This substitution is necessary since only a limited number of ground truth boxes are annotated with 3D keypoints labels. By employing this approach, we simplify and expedite the training process. Additionally, inspired by \cite{zheng2022multi}, we introduce a point-wise segmentation task for keypoints as an auxiliary task to improve the performance of 3D keypoints regression. The pseudo segmentation labels $Y_{kpseg} \in \mathbb{R}^{N_{max} \times (N_{kp} + 1)}$ are generated by assigning each 3D keypoint's type to its top $K$ nearest points. This auxiliary task is supervised using cross-entropy loss, expressed as $\mathcal{L}_{kpseg}$.

To facilitate the 3D keypoints regression, we divide it into two branches: one for the regression over the X and Y axes and another for the regression over the Z axis. This division is based on our observation that predicting the offset along the Z axis is comparatively easier than predicting it along the X and Y axes. We employ a smooth L1 loss to supervise these regression branches, denoting them as $\mathcal{L}_{xy}$ and $\mathcal{L}_{z}$. Note that only the visible 3D keypoints contribute to the regression losses. In addition, we treat the visibility of the keypoints as a binary classification problem. We supervise it using binary cross-entropy loss as $\mathcal{L}_{vis}$.

Our first stage of LidarMultiNet is pretrained following the instructions in \cite{ye2022lidarmultinet} and frozen during the 3D keypoints' training phase. We introduce weight factors for each loss component, and our final loss function is formulated as follows:
\begin{equation}
    \mathcal{L}_{total} = \lambda_{1} \mathcal{L}_{xy} + \lambda_{2} \mathcal{L}_{z} + \lambda_{3} \mathcal{L}_{vis} + \lambda_{4} \mathcal{L}_{kpseg}
\end{equation}
where $\lambda_{1}$, $\lambda_{2}$, $\lambda_{3}$, $\lambda_{4}$ are weight factors and fixed at values of 5, 1, 1, and 1, respectively.

%% file: experiment.tex
\section{Experiments}

\subsection{Dataset}

Waymo Open Dataset (WOD) released the human keypoint annotation on the v1.3.2 dataset that contains LiDAR range images and associated camera images. We use v1.4.2 for training and validation. The 14 classes of keypoints for evaluation are defined as nose, left shoulder, left elbow, left wrist, left hip, left knee, left ankle, right shoulder, right elbow, right wrist, right hip, right knee, right ankle, and head center. In comparing with methods that rely also on images, it is worth noting that there are 144709 objects with 2D keypoint annotations, while only 8125 objects with 3D keypoint annotations are available for training. 

\subsection{Metrics}
We use mean per-joint position error (MPJPE) and Pose Estimation Metric (PEM) as the metrics to evaluate our method. In MPJPE, the visibility of predicted joint $i$ of one human keypoint set $j$  is represented by $v_{i}^{j}\in[0, 1]$, indicating whether there is a ground truth for it. As such, the MPJPE over the whole dataset is:
\begin{equation}
    \textbf{MPJPE}(Y,\hat{Y}) = \frac {1}{\sum_{i,j}v_ {i}^ {j}}  \sum_{i,j} v_{i}^{j} ||Y_{i}^{j} - \hat{Y}_{i}^{j} ||_{2}, 
\end{equation}
where $Y$ and $\hat{Y}$ are the ground truth and predicted 3D coordinates of keypoints.

PEM is a new metric created specifically for the WOD Pose Estimation challenge. Besides keypoint localization error and visibility classification accuracy, it is also sensitive to the rates of false positive and negative object detections, while remaining insensitive to the Intersection over Union (IoU) of object detection. PEM is calculated as a weighted sum of the MPJPE over visible matched keypoints and a penalty for unmatched keypoints, as follows:
\begin{equation}
\textbf{PEM}(Y,\hat{Y}) = \frac{\sum_{i\in M}\left\|y_{i} - \hat{y}_{i}\right\|_2 + C|U|}{|M| + |U|},
\end{equation}
where $M$ is the set of indices of matched keypoints, $U$ is the set of indices of unmatched keypoints, and $C=0.25$ is a constant penalty for unmatched keypoints. Given the partial labeling of WOD keypoints data, MPJPE evaluates keypoint localization accuracy only for matched keypoints, whereas PEM measures both keypoint localization accuracy and object detection quality, better aligning with real-world scenarios.

\begin{table*}[t]
\begin{minipage}{1\textwidth}
\centering
\caption{PEM and MPJPE results on the \texttt{test} split of WOD.}
\label{tab:WOD_pose_test}
\resizebox{\linewidth}{!}{
\begin{tabular}{l|cc|cc|cc|cc|cc|cc|cc|cc}
\hline
& \multicolumn{2}{c}{shoulders} &\multicolumn{2}{c}{elbows}& \multicolumn{2}{c}{wrists} & \multicolumn{2}{c}{hips} & \multicolumn{2}{c}{knees} & \multicolumn{2}{c}{ankles}& \multicolumn{2}{c}{head} &\multicolumn{2}{c}{all}\\
    Model & PEM & MPJPE & PEM & MPJPE& PEM & MPJPE& PEM & MPJPE& PEM & MPJPE& PEM & MPJPE& PEM & MPJPE& PEM & MPJPE  \\
    \hline
    baseline  & 0.2323 & 0.1894 & 0.2354 & 0.2083 & 0.2391 & 0.2240 & 0.2334 & 0.1807 & 0.2327 & 0.1934 & 0.2345 & 0.2250 & 0.2376 & 0.1984 & 0.2349 & 0.2022\\
    KTD & 0.2261 & 0.1876 & 0.2301 & 0.2065 & 0.2349 & 0.2227 & 0.2276 & 0.1790 & 0.2267 & 0.1919 & 0.2290 & 0.2237 & 0.2328 & 0.1973 & 0.2295 & 0.2007\\
    \textbf{\kptr~} &0.1428 & 0.0462 & 0.1511 & 0.0578 & 0.1771 & 0.0951 & 0.1519 & 0.0562 & 0.1477 & 0.0578 & 0.1479 & 0.0663 & 0.1544 & 0.0443 & 0.1524 & 0.0594\\
    \hline
\end{tabular}
}
\end{minipage}
\end{table*}

\subsection{Implementation Details}
Throughout all our experiments, we use a pretrained LidarMultiNet \cite{ye2022lidarmultinet} as the first stage of our framework, which remains frozen during the training phase of the second stage. For additional network and training specifics regarding our first stage, please refer to LidarMultiNet \cite{ye2022lidarmultinet}.

Regarding KPTR, the dimensions of the inputs, namely $C_{point}$, $C_{voxel}$, and $C_{BEV}$, are set to 3, 32, and 512, respectively. The size of the compressed features, denoted as $C_{compressed}$, is 32. We cap the maximum number of points per bounding box at 1024. For the transformer architecture, similar to the recent work \cite{zanfir2023hum3dil}, we utilize $L = 4$ stages, an embedding size of $C_{tr} = 256$, a feed-forward network with internal channels of 256, and 8 heads for the MultiHeadAttention layer. The total number of 3D keypoints $N_{kp}$ is 14. After calculation, the parameter size of KPTR is only 9.61M, accounting for 22.4\% of the total parameters.

During training, we incorporate various data augmentations, including standard random flipping, global scaling, rotation, and translation.  It is important to note that flipping the point clouds has an impact on the relationships between the 3D keypoints annotations, similar to the mirror effect. When performing a flip over the X-axis or Y-axis, the left parts of the 3D keypoints are exchanged with the right parts of the 3D keypoints.

To train our model, we use the AdamW optimizer along with the one-cycle learning rate scheduler for a total of 20 epochs. The training process utilizes a maximum learning rate of 3e-3, a weight decay of 0.01, and a momentum ranging from 0.85 to 0.95. All experiments are conducted on 8 Nvidia A100 GPUs, with a batch size set to 16.

\begin{table}[t]
\centering
\caption{The comparison on the WOD \texttt{val} split. $\ast$: reported by~\cite{zanfir2023hum3dil}, where the result is tested on randomly selected 50\% of subjects from the WOD \texttt{val} split. ``L", ``CL" denote LiDAR-only, camera \& LiDAR fusion methods.}
\label{table:cmp}
\resizebox{\linewidth}{!}{
\begin{tabular}{l|c|c|c}
\hline
Method & Modal& gt box & MPJPE(cm)$\downarrow$ \\ \hline
ContextPose~\cite{ma2021context} $\ast$ & C & $\checkmark$ &  10.82 \\
Multi-modal~\cite{zheng2022multi} & CL &$\checkmark$ &  10.32 \\
GC-KPL~\cite{zheng2022multi} & L &$\checkmark$ &  10.10 \\
THUNDR~\cite{zanfir2021thundr} $\ast$ & C & $\checkmark$ &  9.62 \\
THUNDR~\cite{zanfir2021thundr}  w/ depth $\ast$ & CL & $\checkmark$ & 9.20 \\
Bauer et al.~\cite{bauer2023weakly} & CL &$\checkmark$ &  8.58 \\
HUM3DIL~\cite{zanfir2023hum3dil} $\ast$ & CL & $\checkmark$ &   6.72 \\
\hline
\kptr & L  & &  \textbf{6.16}
 \\\hline
\end{tabular}
}
\end{table}

\subsection{Main Pose Estimation Results}

We tested the performance on the official online test submission site. We trained our model using the combined dataset of Waymo's training and validation splits. The results, presented in Table \ref{tab:WOD_pose_test}, demonstrate the impressive performance of our \kptr, achieving a PEM of 0.1524, an MPJPE of 0.0594, and ranking \nth{1} place on the leaderboard. Notably, our \kptr~outperforms all other methods across all categories in terms of both PEM and MPJPE.

To conduct a comprehensive performance analysis of our \kptr, we compare it with other SOTA methods on the validation set, as shown in Table \ref{table:cmp}. It is important to note that all previous methods were evaluated on a subset of the WOD validation split. Additionally, these methods simplify the problem by providing ground truth 3D bounding boxes along with associated ground truth 2D bounding boxes as inputs during testing. Despite some of these methods incorporating camera and LiDAR fusion or 2D weakly supervision, our \kptr~outperforms them all in terms of MPJPE, achieving an impressive MPJPE of 6.16cm.

\begin{table}[t]
\centering
\caption{The ablation of improvement of each component on the WOD \texttt{val} split.}
\label{table:ablation}
\resizebox{\linewidth}{!}{
\begin{tabular}{c|cccc|cc}
\hline
Baseline & 2nd & seg aux & transformer & box feat & PEM$\downarrow$ & MPJPE $\downarrow$\\ \hline
$\checkmark$ &              &              &              & & 0.1908   &  0.1801 \\
$\checkmark$ & $\checkmark$ &              &              & & 0.1176   &  0.0865 \\
$\checkmark$ & $\checkmark$ & $\checkmark$ &              & & 0.1149   &  0.083  \\
$\checkmark$ & $\checkmark$ & $\checkmark$ & $\checkmark$ & &0.1044  &  0.0703 \\
$\checkmark$ & $\checkmark$ & $\checkmark$ & $\checkmark$ &$\checkmark$ &\textbf{0.0976} &  \textbf{0.0616} 
 \\\hline
\end{tabular}
}
\end{table}

\begin{table}[t]
\centering
\caption{The ablation of training with GT boxes or predicted boxes on the WOD \texttt{val} split. ``gt box'' denotes training with GT boxes only, ``pred box'' denotes training with predicted boxes, ``multiple box'' denotes assigning multiple predicted boxes to each GT through their IoUs, and ``add gt'' denotes training with predicted boxes and GT boxes.}
\label{table:ablation_pred_gt}
\resizebox{\linewidth}{!}{
\begin{tabular}{c|ccc|cc}
\hline
gt box & pred box & multiple box & add gt & PEM$\downarrow$ & MPJPE $\downarrow$\\ \hline
$\checkmark$ &              &              &              &\textbf{0.0976} &  \textbf{0.0616} \\
 & $\checkmark$ &              &              & 0.1009   &  0.0659 \\
 & $\checkmark$ & $\checkmark$ &              & 0.1005   &  0.0645  \\
 & $\checkmark$ & $\checkmark$ & $\checkmark$ & 0.0995  &  0.0633 \\
 \hline
\end{tabular}
}
\end{table}

\begin{table}[t]
\centering
\caption{The ablation of point features selections on the WOD \texttt{val} split.}
\label{table:ablation_feat}
\resizebox{\linewidth}{!}{
\begin{tabular}{cc|cc}
\hline
point features & voxel features & PEM$\downarrow$ & MPJPE $\downarrow$\\ \hline
 $\checkmark$ & $\checkmark$ &\textbf{0.0976} &  \textbf{0.0616} \\
 $\checkmark$ & & 0.0992   &  0.0639  \\
  & $\checkmark$ & 0.1003   &  0.0652 \\
 \hline
\end{tabular}
}
\end{table}

\begin{figure*}
    \centering
    \begin{subfigure}{0.47\textwidth}
        \includegraphics[width=1.0\linewidth]{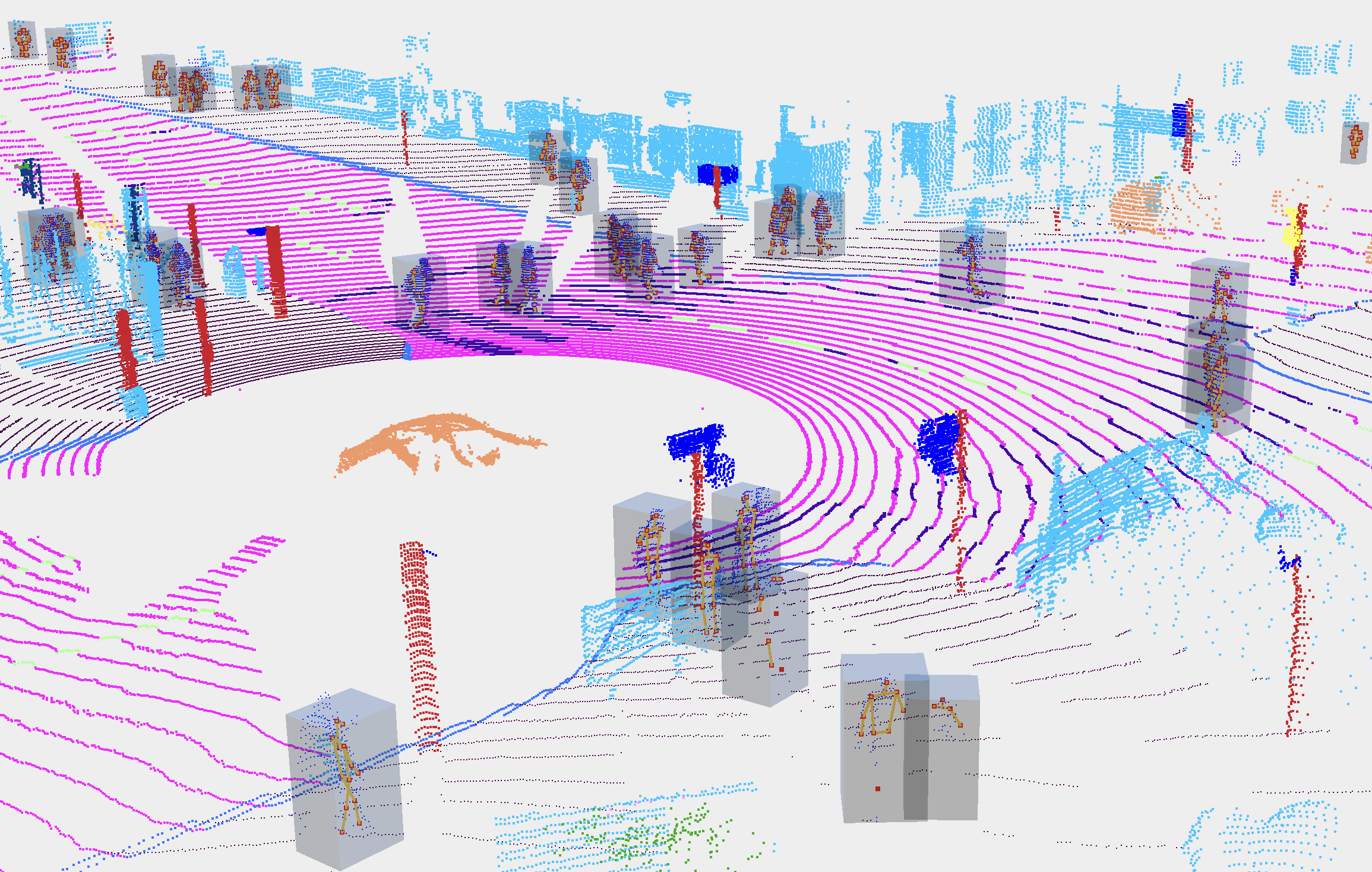}
    \end{subfigure}
    \begin{subfigure}{0.515\textwidth}
        \includegraphics[width=1.0\linewidth]{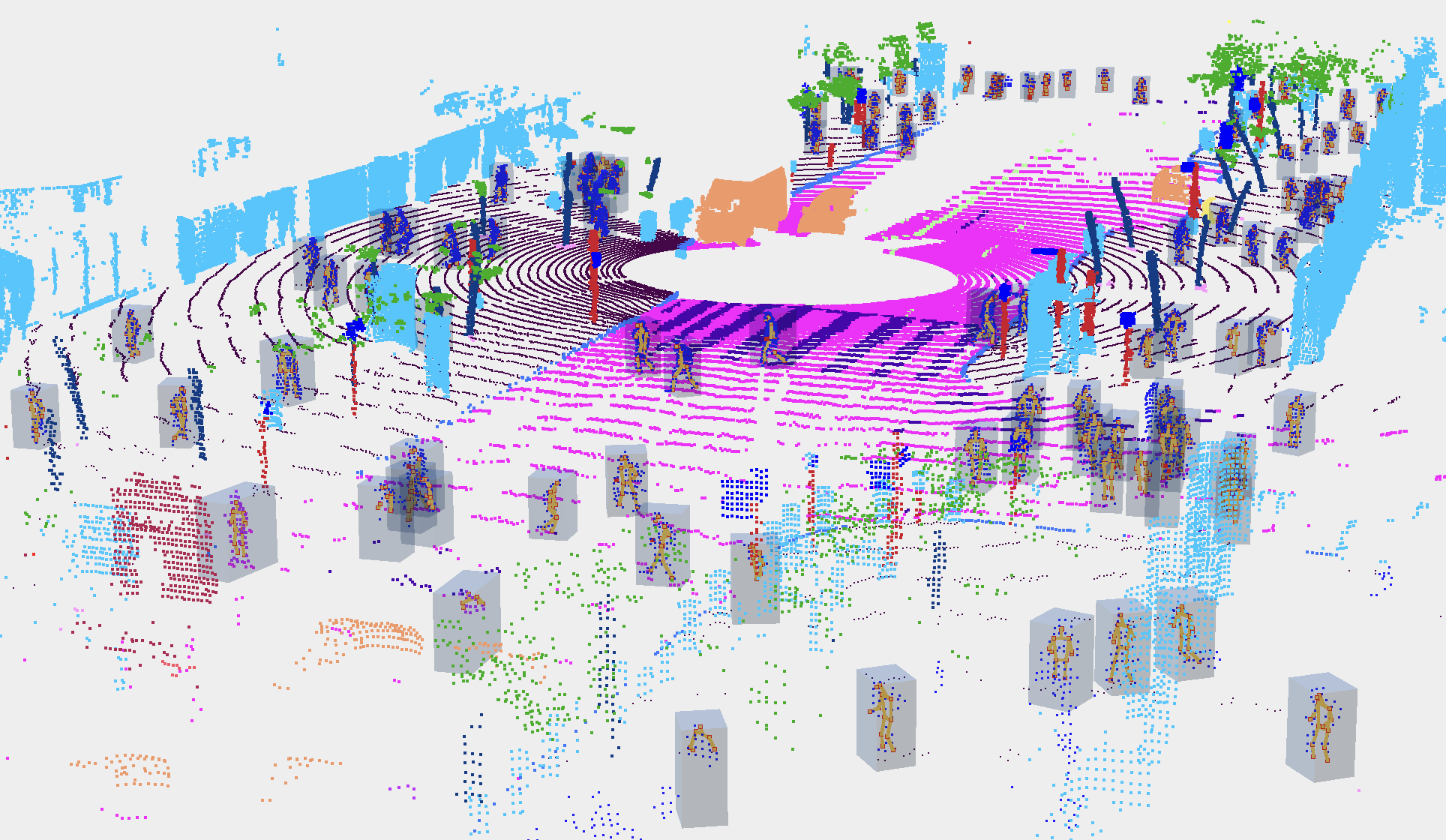}
    \end{subfigure}
    \caption{Prediction results on the whole scene with a significant number of pedestrians in the validation set.}
    \label{fig:whole_scene}
\end{figure*}

\subsection{Ablation Study}

Table \ref{table:ablation} shows a comparison of the performance between the first stage and \kptr, as well as the contribution of each component in the second stage to the overall performance. The first stage results are directly output from the center head following the BEV feature map. Given that the BEV feature map is primarily driven by the detection task and has low resolution, it lacks fine-grained features, resulting in mediocre performance. The second stage which is similar to second-stage refinement module in LidarMultiNet \cite{ye2022lidarmultinet}, however, significantly improves upon introducing point-wise fine-grained features. Further enhancements are achieved by adding the keypoint segmentation auxiliary task, employing the transformer structure, and incorporating box features, all of which contribute to varying degrees to performance improvement of the model.

In Table \ref{table:ablation_pred_gt}, our results show that integrating ground truth bounding boxes into the training process not only proves to be efficient but also improves model performance. Training with only predicted bounding boxes with the highest IoU with their associated ground truth bounding boxes results in a performance drop. Allowing multiple predicted bounding boxes matched with each ground truth one improves the performance; further improvements are noted when these predicted boxes are combined with ground truth bounding boxes. However, using ground truth bounding boxes exclusively still achieves the best performance.

Table \ref{table:ablation_feat} presents an ablation study of point features and voxel features. Here, the point features refer to the additional features like intensity, elongation, timestamp, and so on, while the $xyz$ coordinates of the points are retained. Within the KPTR framework, we observe that the performance is optimal when both point and voxel features are combined. Removing either feature results in a decline in performance.

\subsection{Qualitative Results}

\begin{figure}
    \centering
    \begin{subfigure}{0.325\linewidth}
        \includegraphics[width=1.0\linewidth]{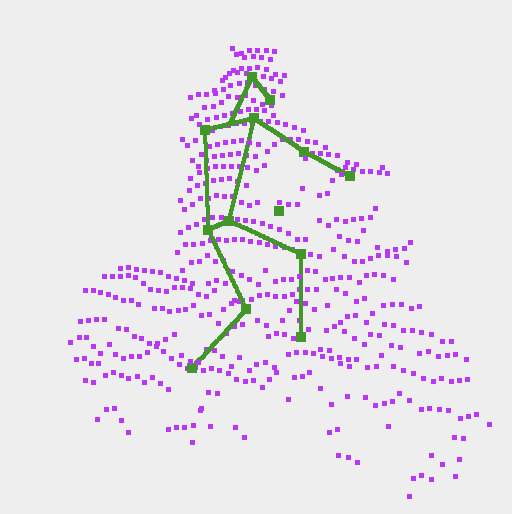}
    \end{subfigure}
    \hfill
    \begin{subfigure}{0.325\linewidth}
        \includegraphics[width=1.0\linewidth]{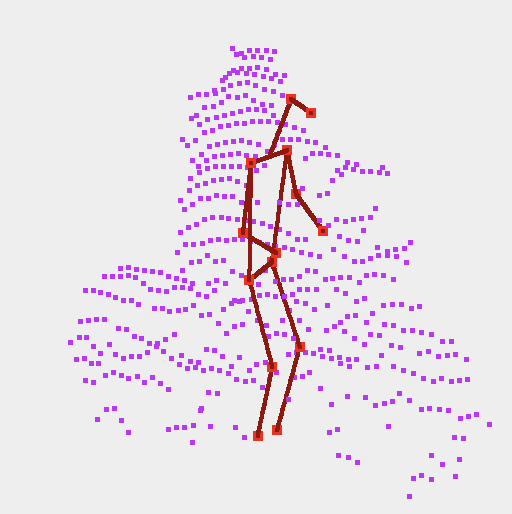}
    \end{subfigure}
    \hfill
    \begin{subfigure}{0.325\linewidth}
        \includegraphics[width=1.0\linewidth]{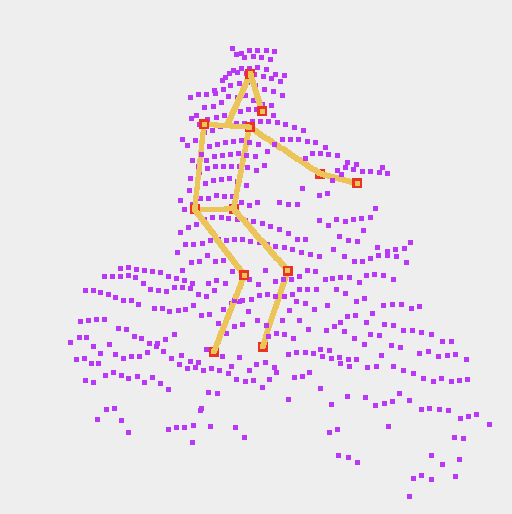}
    \end{subfigure}
    
    \begin{subfigure}{0.325\linewidth}
        \includegraphics[width=1.0\linewidth]{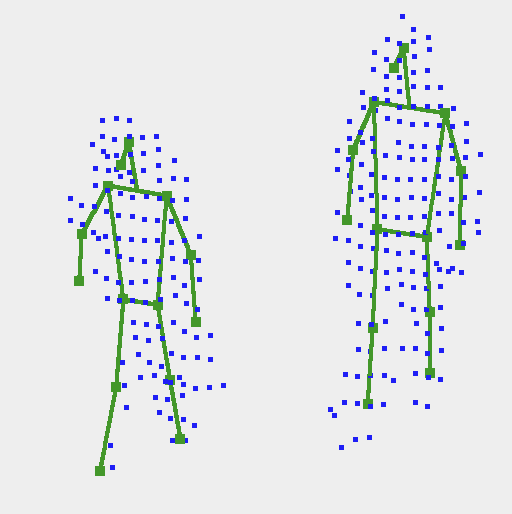}
    \end{subfigure}
    \hfill
    \begin{subfigure}{0.325\linewidth}
        \includegraphics[width=1.0\linewidth]{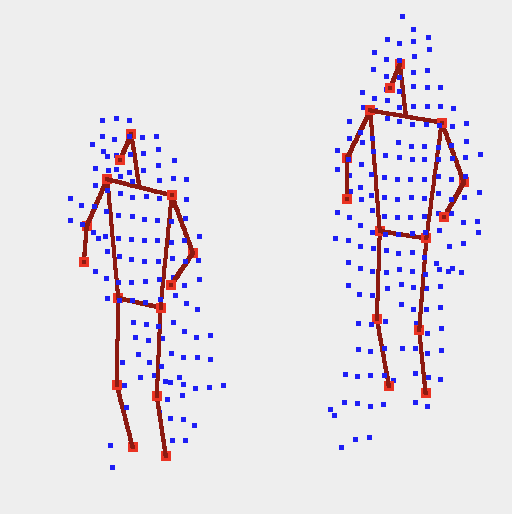}
    \end{subfigure}
    \hfill
    \begin{subfigure}{0.325\linewidth}
        \includegraphics[width=1.0\linewidth]{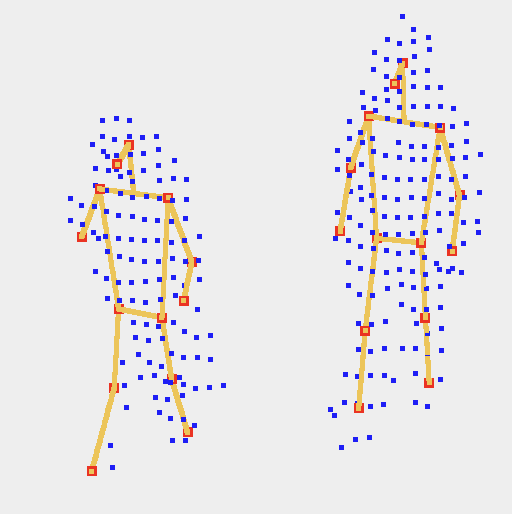}
    \end{subfigure}

    \begin{subfigure}{0.325\linewidth}
        \includegraphics[width=1.0\linewidth]{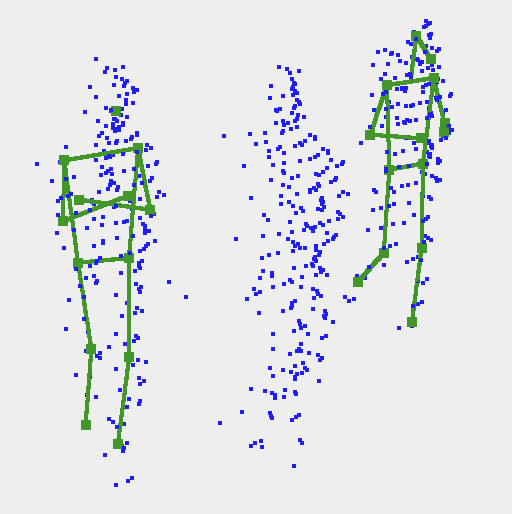}
        \caption{Ground Truth}
    \end{subfigure}
    \hfill
    \begin{subfigure}{0.325\linewidth}
        \includegraphics[width=1.0\linewidth]{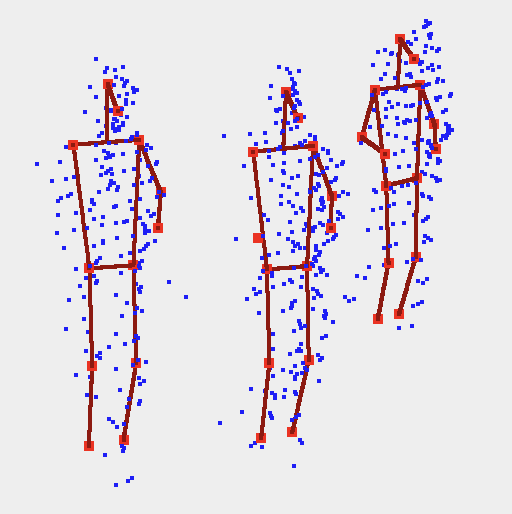}
        \caption{1st stage}
    \end{subfigure}
    \hfill
    \begin{subfigure}{0.325\linewidth}
        \includegraphics[width=1.0\linewidth]{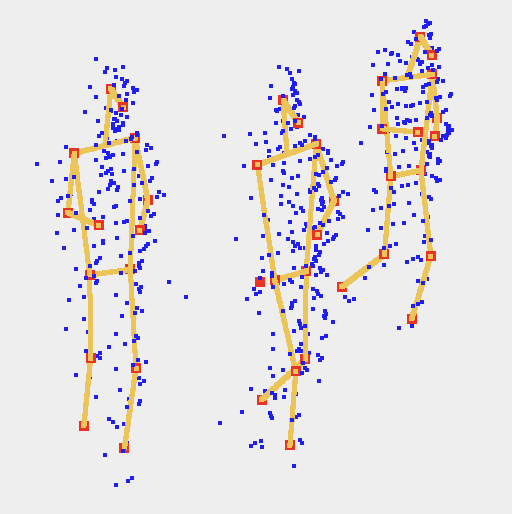}
        \caption{\kptr}
    \end{subfigure}
    \caption{Prediction results compared to the Ground Truth and the 1st stage results.}
    \label{fig:cmp_cases}
\end{figure}

Figure \ref{fig:opening} shows the output predictions of our model for one frame in the validation set, viewed from a particular angle. The input is solely 3D LiDAR point cloud. Remarkably, our network simultaneously outputs results of 3D semantic segmentation, 3D bounding boxes, as well as their 3D keypoints (red) along with the corresponding wireframes (yellow) for visualization. Our model also predicts visibility, for example, the left knee of the second person from the left is predicted as invisible, while the left foot is visible. Both feet of the third person from the right are predicted as invisible. The right elbow of the sixth person from the right is predicted as invisible, however, the right hand is visible. Figure \ref{fig:whole_scene} demonstrates some prediction results on the whole scene in the validation set.
 
Figure \ref{fig:cmp_cases} presents a selection of predictions made on the validation set. From left to right, the three columns represent ground truths, the predictions of the 1st stage, and the predictions of \kptr, respectively. Each row showcases the same group of objects. As can be observed, across all three groups, the performance of \kptr~noticeably surpasses that of the 1st stage output. The first row highlights a cyclist for whom ground truth annotations are extremely limited. Despite the limited amount of annotations, \kptr~still manages to deliver meaningful output. In the second row, \kptr~is strikingly close to the ground truth, with the exception of an FN visibility for the right hand of the pedestrian on the left. The third row demonstrates that even on the pedestrian without ground truth annotations, \kptr~still produces satisfactory results. For the running pedestrian on the right, \kptr~performs very well. However, the left pedestrian's head center is an FP case, and the crossed hands pose is a difficult case given the small amount of similar ground truth annotations available.